\documentclass[sn-mathphys-num]{sn-jnl}

\usepackage{graphicx}%
\usepackage{multirow}%
\usepackage{amsmath,amssymb,amsfonts}%
\usepackage{amsthm}%
\usepackage{mathrsfs}%
\usepackage[title]{appendix}%
\usepackage{xcolor}%
\usepackage{textcomp}%
\usepackage{manyfoot}%
\usepackage{booktabs}%
\usepackage{algorithm}%
\usepackage{algorithmicx}%
\usepackage{algpseudocode}%
\usepackage{listings}%

\theoremstyle{thmstyleone}%
\theoremstyle{thmstyletwo}%

\theoremstyle{thmstylethree}%

\begin{document}

\title[Article Title]{EFA-YOLO: An Efficient Feature Attention Model for Fire and Flame Detection}


\author*[1]{\fnm{Weichao} \sur{Pan}}\email{202211107025@stu.sdjzu.edu.cn}

\author[1]{\fnm{Xu} \sur{Wang}}\email{202311102025@stu.sdjzu.edu.cn}

\author[2]{\fnm{Wenqing} \sur{Huan}}\email{202212105033@stu.sdjzu.edu.cn}

\affil*[1]{\orgdiv{School of Computer Science and Technology}, \orgname{Shandong Jianzhu University}, \orgaddress{\street{Ganggou Street}, \city{Jinan}, \postcode{250000}, \state{Shandong}, \country{China}}}

\affil[2]{\orgdiv{School of Science}, \orgname{Shandong Jianzhu University}, \orgaddress{\street{Ganggou Street}, \city{Jinan}, \postcode{250000}, \state{Shandong}, \country{China}}}


\abstract{As a natural disaster with high suddenness and great destructiveness, fire has long posed a major threat to human society and ecological environment. In recent years, with the rapid development of smart city and Internet of Things (IoT) technologies, fire detection systems based on deep learning have gradually become a key means to cope with fire hazards. However, existing fire detection models still have many challenges in terms of detection accuracy and real-time performance in complex contexts. To address these issues, we propose two key modules: EAConv (Efficient Attention Convolution) and EADown (Efficient Attention Downsampling). The EAConv module significantly improves the feature extraction efficiency by combining an efficient attention mechanism with depth-separable convolution, while the EADown module enhances the accuracy and efficiency of feature downsampling by utilizing spatial and channel attention mechanisms in combination with pooling operations. Based on these two modules, we design an efficient and lightweight flame detection model, EFA-YOLO (Efficient Feature Attention YOLO). Experimental results show that EFA-YOLO has a model parameter quantity of only 1.4M, GFLOPs of 4.6, and the inference time per image on the CPU is only 22.19 ms. Compared with existing mainstream models (e.g., YOLOv5, YOLOv8, YOLOv9, and YOLOv10), EFA-YOLO exhibits a significant enhancement in detection accuracy (mAP) and inference speed, with model parameter amount is reduced by 94.6\% and the inference speed is improved by 88 times.}




\maketitle

\section{Introduction}\label{sec1}

As a kind of sudden and extremely destructive disaster \cite{1,2,3}, fire has long posed a serious threat to human society and the natural environment. With the acceleration of urbanization and industrialization, the complexity of building structures and the density of crowds have gradually increased, making the frequency and hazard of fire rise \cite{4,5,6,7}. According to the statistics of the International Fire Protection Organization (IFPO), millions of fire accidents occur globally every year, which cause a large amount of loss of life and property and irreversible damage to the ecological environment. In forested areas, the rapid spread of fire often leads to the destruction of large areas of forest resources, which in turn exacerbates ecological problems such as soil erosion and air pollution.

The diversity and complexity of fire hazards make fire prevention and control particularly important. Fire hazards can come from a variety of sources such as building structures, electronic equipment, improper storage and handling of flammable materials, and are often hidden and sudden. Traditional fire detection techniques rely on devices such as smoke sensors and temperature alarms, and although these devices are able to detect the occurrence of fire to a certain extent, their response efficiency and accuracy are often limited in open spaces, outdoor environments, or in the early stages of a fire \cite{7,8,9}. Especially in the early stages of a fire, flames are small and not easily captured by conventional sensors, which increases the difficulty of preventing and controlling fire hazards.

Therefore, the development of novel fire detection technologies, especially intelligent fire detection systems based on image processing and deep learning \cite{10,11,12}, is of extreme practical importance. These technologies can utilize the visual characteristics of flames to achieve early fire warning and reduce the hazards caused by fires through precise location and fast response. With the popularization of smart city and Internet of Things (IoT) technologies, vision-based fire detection systems will provide more effective solutions for the monitoring and management of modern fire hazards.

In recent years, with the increasing demand for fire prevention and control, the research on fire detection in different scenarios has gradually deepened. Researchers have proposed a variety of improvement methods to address the limitations of the existing fire detection techniques, especially on the problems of flame detection in complex backgrounds, urban fire monitoring with high real-time requirements, and flame detection of small targets, and a variety of innovative algorithms have emerged. To solve these problems, researchers have not only optimized the structure of the detection model, but also adopted technical means such as data enhancement, multi-scale feature extraction and attention mechanism. The work and contributions of some researchers in fire detection are detailed below \cite{13,14,15,16,17,18,19,20,21,22,23,24,25,26}.

Wang et al. \cite{13} proposed the YOLOv5s-ACE algorithm to address the problems of low detection accuracy, slow detection and rough feature extraction in the context of complex forest fires. The algorithm first extends the small object sample set by Copy-Pasting data enhancement to reduce the risk of overfitting during model training. Secondly, an empty space pyramid pooling (ASPP) module is chosen to replace the SPP module in YOLOv5, which expands the sensory field and improves the accurate localization of small-target forest flames. Finally, the Convolutional Block Attention Module (CBAM) is added to further filter the key features and reduce the background interference. Sun et al. \cite{14} proposed the AERNet model to address the problem of the high false detection rate of existing fire detectors in the multi-scale variations of flames and smoke, as well as in the complex background. The model employs a lightweight backbone network, Squeeze-and-Excitation GhostNet, to reduce the number of model parameters and enhance the feature extraction capability. In addition, by constructing a multi-scale detection module, the contributions of different features are selectively emphasized both spatially and channel-wise, which improves the detection accuracy and speed. Cao et al. \cite{15} proposed a novel detection technique based on the improved YOLOv5 model to address the failure problem of the existing fire detection methods in detecting small or hidden fires. The efficiency of the YOLOv5 model in feature extraction is improved by incorporating a global attention mechanism and a reparameterized convolution module, and a bidirectional feature pyramid network (BiFPN) is used for feature information fusion, which improves the processing of local information. Yang et al. \cite{16} proposed a YOLOv5s network based on a real-time image processing requirement in urban fire monitoring. based lightweight detection model. The model introduces the Squeeze-and-Excitation module for image filtering and classification to meet the demand for rapid data screening in smart city fire monitoring systems. Yu et al. \cite{17} proposed a deep learning-based early fire detection algorithm to solve the problem of high false alarm rate of traditional smoke alarms. The algorithm combines a smoke detector, a thermal imaging camera, and a YOLOv7 model, and ultimately achieves accurate detection of actual fires by eliminating the bounding box of non-fire reports through a deep learning model. Lv et al. \cite{18} proposed a lightweight fire detection algorithm based on YOLOv5s to address the challenge of the fire detection algorithm's low recognition rate of small fire targets in complex environments. By introducing the CoT (Contextual Transformer) structure as well as the CSP1\_CoT module in the backbone network, the number of parameters of the model is effectively reduced while the detection ability of small targets is improved. Wang et al. \cite{19} proposed the DATFNets framework for the limitations of fire detection in complex contexts and optimized the network performance through a dynamic adaptive allocation strategy and the weighted loss function to optimize the network performance. Cao et al. \cite{20} proposed a fire detection method based on feature fusion and channel attention for the detection of small target flames in complex scenes. The method enhances the feature extraction capability by using deformable convolution in the backbone network, and improves the localization capability of small fire targets through the channel attention mechanism. Chen et al. \cite{21} proposed the GS-YOLOv5 model, which adopts the Super-SPPF structure and the C3Ghost module to effectively reduce the number of model parameters, and introduces the coordinate attention (CA) module to improve the detection accuracy. Wu et al. \cite{22} proposed a multi-scale fire detection method combining CNN and Transformer, using a CNN module in shallow feature extraction and a Transformer module in deep feature extraction for global sensing. Li et al. \cite{23}, in response to the limitations of traditional fire detectors that are difficult to effectively detect small fires in the early stage, proposed a early fire detection system based on shallow guided depth network (SGDNet). The system first extracts flame features in YCbCr color space, and then fuses the shallow and deep features through the SGD module. The model is optimized by a redesigned backbone, detection head and IoU, which enables efficient detection on embedded devices. Dou et al. \cite{24} proposed an image-based non-contact fire detection technique for contact fire sensors that are susceptible to interference from non-fire particles. Their study demonstrated the advantages of YOLOv5 in mAP and FPS by comparing eight existing object detection models, and further optimized YOLOv5s network by introducing a CBAM module, BiFPN structure, and inverse convolution, which significantly improves the detection accuracy and processing speed of the model. Wang et al. \cite{25} In response to the YOLOv7's recognition of small, dense fire-smoke targets with its limitations of YOLOv7 in recognizing small dense fire and smoke targets, the FS-YOLO model was proposed. The model reduces local feature dependency by enhancing the Swin Transformer module and introduces an efficient channel attention mechanism to reduce false alarms in fire detection. In addition, the study developed a dual dataset containing real fire scenes and fire and smoke images to simulate complex conditions such as occlusion and lens blurring. Wang et al. \cite{26} proposed an improved YOLOX multiscale fire detection method to address the problem that the traditional fire detection methods are ineffective when the range of flame and smoke targets is extended. The method reduces the information loss of high-level feature maps and enhances the feature representation capability by designing a novel feature pyramid model (HC-FPN). In addition, a small target data enhancement strategy is used to extend the forest fire dataset so that the model is more adapted to real forest fire scenarios.

Although researchers have improved the detection accuracy and speed by introducing the attention mechanism, lightweight backbone network, and multi-scale feature extraction, there are still problems such as large number of model parameters and high computational complexity when fire monitoring with high real-time requirements. To cope with these problems, we propose an innovative flame detection model EFA-YOLO (Efficient Feature Attention YOLO). The model realizes efficient feature extraction and downsampling through two key modules: EAConv (Efficient Attention Convolution) and EADown (Efficient Attention Downsampling). The EAConv module combines the efficient attention mechanism and deep separable convolution to EADown module enhances the accuracy and efficiency of feature downsampling by fusing spatial and channel attention mechanisms and pooling operations.

Overall, the contributions of this paper are as follows:
\begin{itemize}
 \item \textbf{Two key modules are proposed- EAConv and EADown:} The EAConv module dramatically improves the efficiency and accuracy of feature extraction by means of an efficient attention mechanism and depth-separable convolution. the EADown module combines spatial and channel attention mechanisms and introduces maximal pooling and average pooling operations, which enhance the performance of feature downsampling.
 \item \textbf{Designed and implemented an efficient and lightweight fire flame detection model, EFA-YOLO:} The model significantly reduces the number of parameters and computational complexity of the model while maintaining high detection accuracy, which is especially suitable for real-time fire detection applications.EFA-YOLO improves the detection of complex backgrounds through effective multi-scale feature fusion. Compared with mainstream models (e.g., YOLOv5, YOLOv8, YOLOv9, and YOLOv10), EFA-YOLO exhibits significant improvement in both detection accuracy (mAP) and inference speed. Experimental results show that EFA-YOLO reduces the amount of model parameters by 94.6\% and speeds up the inference time by 88 times, which greatly improves the real-time detection.
 \item \textbf{Provides an effective lightweight solution for embedded devices and smart city scenarios:} Due to its lightweight design, EFA-YOLO is able to operate efficiently on resource-constrained devices, is suitable for fire monitoring systems in smart cities, and provides a flexible and scalable model architecture for future fire detection technologies.
\end{itemize}

\section{Methods}\label{sec2}

In this section, we will provide a comprehensive explanation of the proposed model, providing a detailed description of each module in the network model and clarifying their respective functions. First, an explanation of the overall model will be provided, followed by a detailed explanation of the modules involved, including the EAConv (Efficient Attention Convolution) module, EADown (Efficient Attention Downsampling) module.

\subsection{Overview}

The EFA-YOLO (Efficient Feature Attention YOLO) model is proposed to cope with the multiple challenges faced by current fire detection techniques in complex scenarios, especially the real-time demand in flame detection and the problem of small object detection in complex backgrounds.

\begin{figure}[h]
    \centering
    \includegraphics[width=1\linewidth]{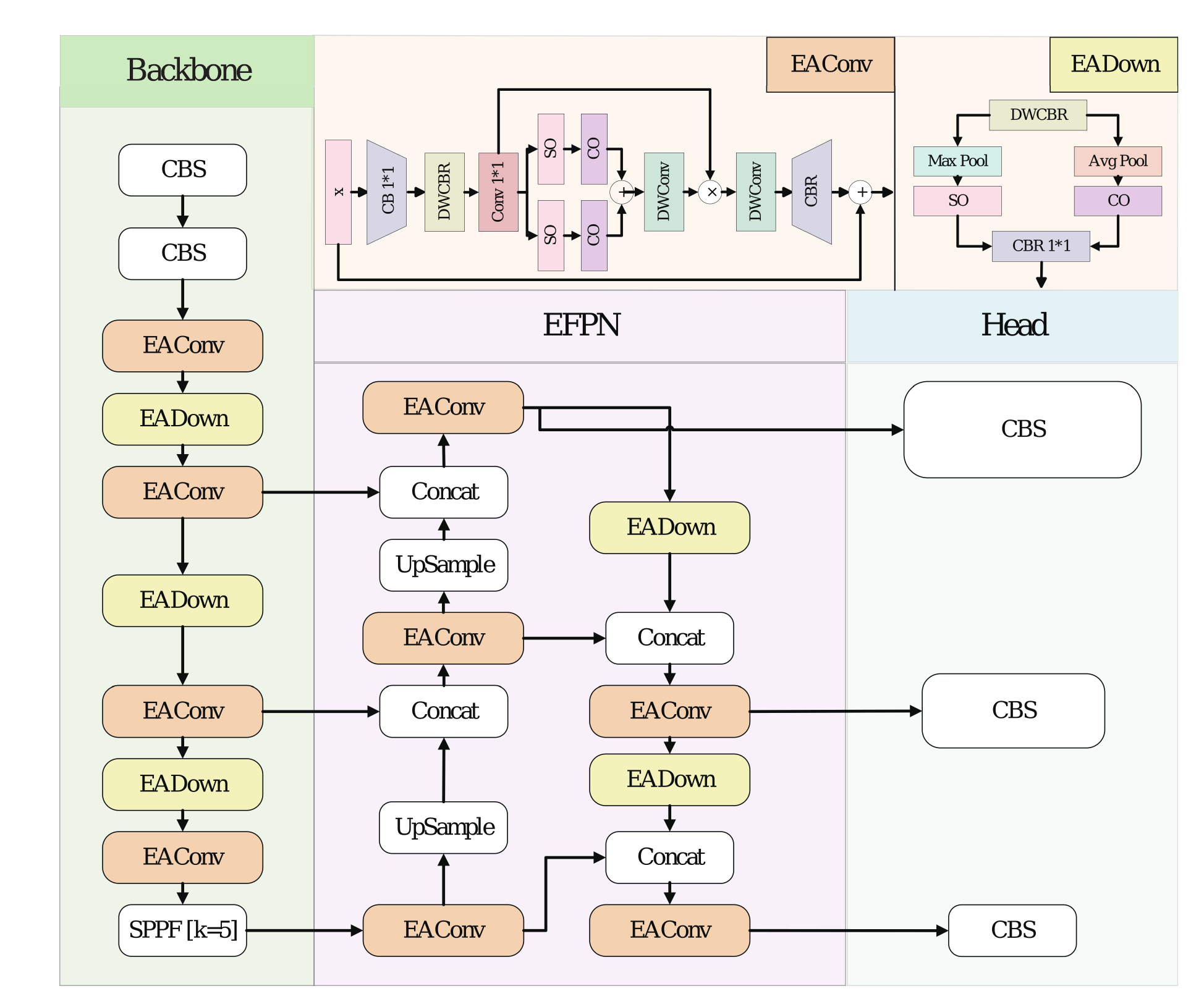}
    \caption{Shows the EFA-YOLO overall framework.}
    \label{fig1}
\end{figure}

The design concept of EFA-YOLO is based on two key modules - EAConv (Efficient Attention Convolution) and EADown (Efficient Attention Downsampling). The EAConv module improves the perception of complex targets such as flames through an efficient feature extraction mechanism, while the EADown module effectively achieves efficient feature downsampling by combining the attention mechanism \cite{27} with pooling operations. The overall model architecture fully considers the multi-scale feature extraction requirements in fire detection, while taking into account the lightweight and computational efficiency of the model. This makes EFA-YOLO not only capable of high-precision detection in complex scenarios, but also significantly shortens the inference time and meets the real-time requirements. Through comprehensive optimization, EFA-YOLO achieves significant improvement in the number of model parameters, inference time, and detection accuracy, which proves that it has a wide range of application prospects in the field of fire detection.

where the SPPF is formulated as follows.

The SPPF module is typically an enhanced version of the SPP (Spatial Pyramid Pooling), which allows different pooling kernel sizes for the same input, followed by concatenation. Here's the formula to explain SPPF:

Let the input feature map be \( X \in \mathbb{R}^{H \times W \times C} \), where \( H \), \( W \), and \( C \) represent the height, width, and channels, respectively.

Steps involved in SPPF:

1. Apply MaxPooling with different kernel sizes:
\begin{equation}
    X_1 = \text{MaxPool}(X, k=5, s=1, p=2)
\end{equation}

Repeat this operation with other pooling layers. For example:
\begin{equation}
    X_2 = \text{MaxPool}(X, k=5, s=1, p=2)
\end{equation}

Similar operations are applied, and the outputs are concatenated:

2. Concatenate the pooled feature maps:
\begin{equation}
    X_{out} = \text{Concat}(X_1, X_2, X_3)
\end{equation}

3. Pass the concatenated feature map through a convolutional layer (often 1x1 convolutions):
\begin{equation}
    X_{out} = \text{Conv}(X_{concat})
\end{equation}
The convolution applies filters to compress and refine the concatenated feature maps.

where the formula for CBS is as follows.

\begin{equation}
    X_{out} = \text{SiLU}(\text{BN}(\text{Conv}(X)))
\end{equation}

\subsection{EAConv}

\begin{figure}[h]
    \centering
    \includegraphics[width=0.7\linewidth]{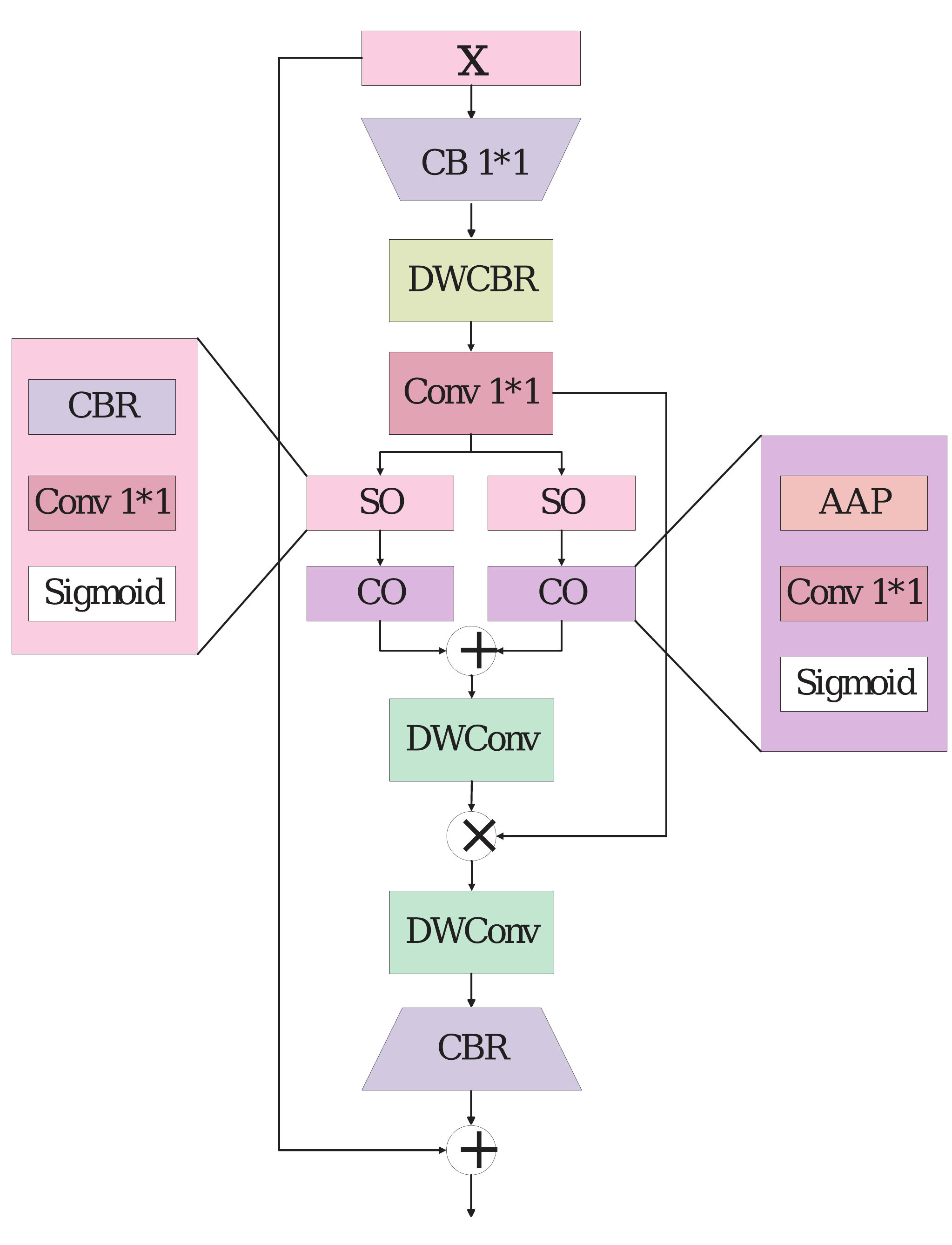}
    \caption{Shows the EAConv module.}
    \label{fig2}
\end{figure}

The EAConv module is one of the cores of EFA-YOLO, which was originally designed to address the limitations of traditional convolution when facing complex scenes and dynamic targets such as flames. Traditional convolutional layers are often unable to accurately capture the spatial and channel information of small targets such as flames when performing feature extraction, which makes the model's detection capability overstretched in complex backgrounds. To this end, the EAConv module significantly improves the accuracy and robustness of flame detection by introducing efficient spatial and channel attention mechanisms that enable the model to intelligently focus on key feature regions in the image. EAConv is shown in Fig~\ref{fig2}.

At the same time, the EAConv module incorporates Depthwise Separable Convolution \cite{28}, a convolution technique that can significantly reduce the amount of computation while maintaining the effectiveness of feature extraction. In this way, EAConv not only achieves a breakthrough in detection accuracy, but also significantly reduces the computational complexity of the model, allowing it to be more easily deployed on resource-constrained devices, such as edge devices and embedded systems. The EAConv module has been repeatedly used in multiple layers of EFA-YOLO, ensuring that features at every scale are adequately represented and processed, and ultimately EAConv is not only a simple optimization of the convolution module, but also an innovation in feature extraction, which enables EFA-YOLO to maintain high computational efficiency while still possessing strong feature capture capability through the combination of attention mechanism and lightweight convolution. Such a design makes EAConv one of the key factors for the whole model to improve detection accuracy and accelerate inference.

\subsection{EADown}

\begin{figure}[h]
    \centering
    \includegraphics[width=0.5\linewidth]{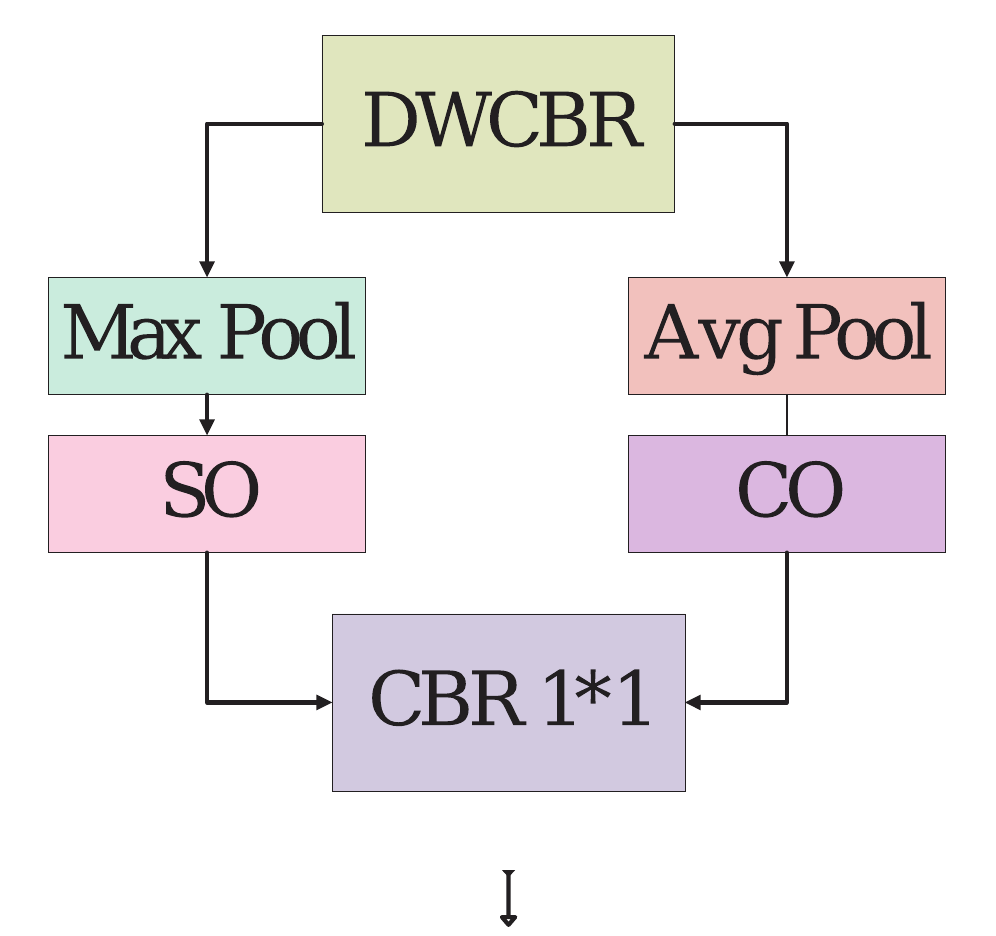}
    \caption{Shows the EADown module.}
    \label{fig3}
\end{figure}

The EADown module focuses on the efficient feature downsampling process, which is another important module in EFA-YOLO. In flame detection scenarios, especially when dealing with small flames or long-distance fires, the model often loses important detail information during downsampling due to the reduced size of the feature map, which can lead to missed or false detections. The EADown module achieves the retention and enhancement of key information during downsampling by introducing spatial attention and channel attention mechanisms combined with the strategies of maximum pooling and average pooling. This design ensures that the model still maintains sensitivity to the flame target during the downsampling phase and effectively prevents feature loss. EADown is shown in Fig~\ref{fig3}.

The EADown module goes beyond a mere downsampling operation and intelligently filters important features through an attention mechanism, thus reducing the feature map size while ensuring that the model maintains sufficient attention to targets such as flames. This allows the model to further extract high-level semantic features while still retaining the perception of low-level features, especially the details of small target flames. In addition, EADown's multi-scale feature extraction capability enhances the model's adaptability to targets of different scales, ensuring that EFA-YOLO performs well when dealing with flame targets of different sizes. With the optimization of the EADown module, EFA-YOLO's performance in dealing with small target flames and complex scenes has been significantly improved. This module complements the EAConv module to ensure that the model remains lightweight while still having strong feature representation and computational efficiency. This enables EFA-YOLO to meet the multiple requirements of real-time, high accuracy and light weight in practical applications, providing a practical solution for fire detection.

\section{Experimental Details}\label{sec3}

\subsection{Dataset}

The fire dataset \cite{29} covers a wide range of fire scenarios, including fires in buildings, grasslands, forests, and different objects such as vehicles (e.g., cars, trucks, motorcycles, and motorized vehicles), with fire sizes ranging from large to small fires. The dataset also distinguishes between day and night, indoor and outdoor fire environments to ensure that the model can cope with different lighting and spatial conditions. In addition, the smoke characteristics in each fire scenario are consistent with the fire, reflecting differences in fire type, size, and environment. In total, there are 2060 labeled datasets, which we randomly divide 8:2 into training and testing sets. This dataset can be used to train models for fire detection and smoke recognition, which can be applied in intelligent scenarios such as fire monitoring and warning systems.

\begin{figure}[h]
    \centering
    \includegraphics[width=1\linewidth]{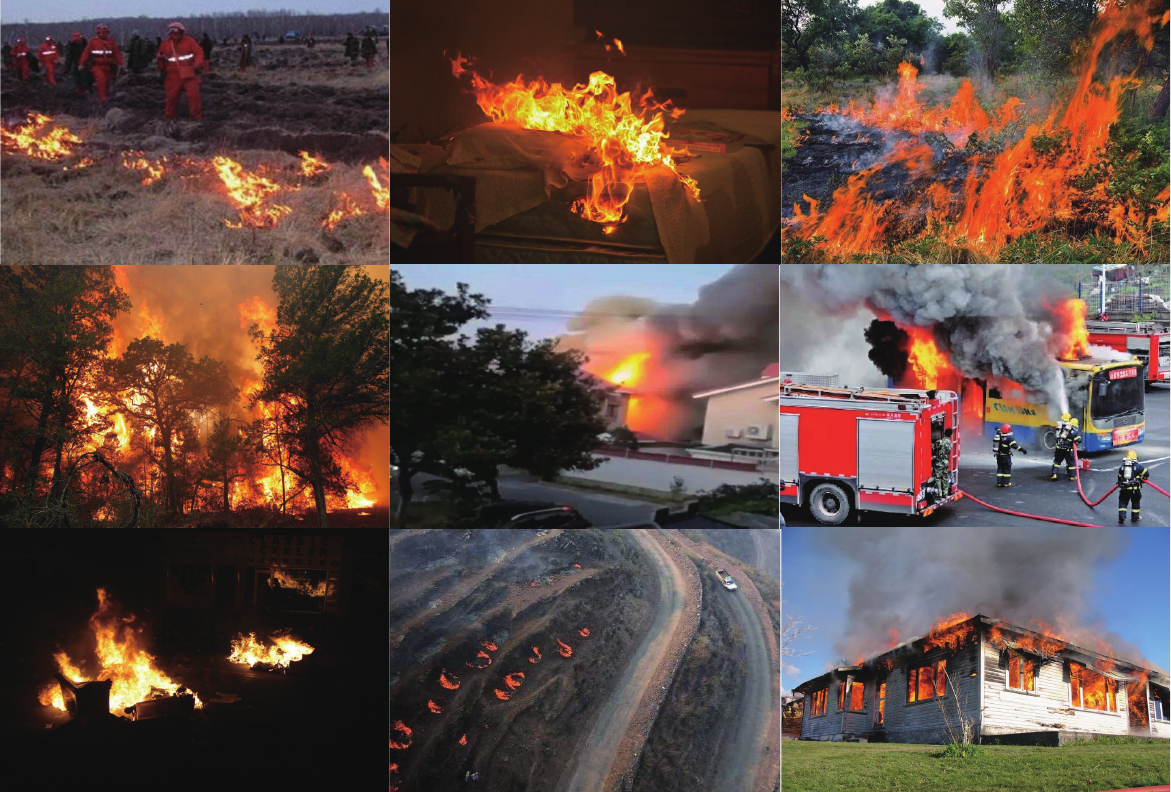}
    \caption{Part of the dataset sample display.}
    \label{fig4}
\end{figure}

\subsection{Experimental environment}

The experimental program was executed on Linux operating system with NVIDIA GeForce RTX 4090D 24G graphics card driver. Pytorch+cu version 11.8 with 2.0.1 was chosen for the deep learning framework, Jupyter Notebook was used for the compiler, Python 3.8 was used as the specified programming language, and all the algorithms used in the comparative analyses were operationally consistent and ran in the same computational settings. The image size was normalized to 640×640×3, batch size was 8, optimizer was SGD, learning rate was set to 0.001, and the number of training periods was 300. An AMD EPYC 9754 processor was used to calculate the inference time.

\subsection{Evaluation metrics}

In this study four key metrics, precision, recall, mAP50 and mAP50:95, were used to assess the performance of the detection model. Precision measures the accuracy of the model in predicting positive categories, while recall assesses the comprehensiveness of the positive categories detected by the model. Whereas, mAP50 and mAP50:95 denote the average accuracy at an IOU threshold of 0.5 and varying from 0.5 to 0.95, respectively, and are used to comprehensively assess the model's detection performance under different conditions \cite{30}.

To further evaluate the complexity and efficiency of the model, the study also uses Params(M) (the number of parameters in the model), GFLOPs (billions of floating-point operations per second), Model Size(MB) (the size of the model), and Inference Time(ms). Params(M) represents the total number of parameters in the model, and the lower the number of parameters, the more lightweight and suitable for deployment in resource-constrained environments. Params(M) represents the total number of parameters in the model, the smaller the number of parameters, the lighter the model is and the more suitable it is to be deployed in resource-constrained environments. GFLOPs represents the computational complexity of the model, the lower the GFLOPs, the lower the computational requirements of the model and the more efficient it is. Model Size(MB) measures the storage requirements of the model, the smaller the model, the easier it is to be deployed in embedded devices. Through the comprehensive evaluation of these metrics, this study effectively measures the performance of the model in terms of detection accuracy, computational complexity and real-time performance, which provides a scientific basis for the optimization and practical application of fire detection models.

\section{Experimental Results and Discussion Analysis}

In order to validate the superior performance of the EFA-YOLO object detection model proposed in this paper, a series of validations are conducted on the above dataset, and several evaluation metrics mentioned above are used for evaluation and analysis. Firstly, this paper introduces the current mainstream object detection models and conducts comparative experiments with the model EFA-YOLO proposed in this paper to prove the superiority of the model proposed in this paper. Then the results of the proposed model in this paper are evaluated, including the analysis of the results of the comparison experiments, the recognition results of the comparison experimental model analysis. Finally, the effectiveness of the modules as well as the structure designed in this paper is verified by ablation experiments.

\subsection{Comparative Experiment}

In order to verify the performance of the proposed model, we compared the EFA-YOLO trained using the training set with YOLOv5 \cite{31}, YOLOv8 \cite{32}, YOLOv9 \cite{33}, and YOLOv10 \cite{34} object detection models. Through this experiment, the superior performance of the model is demonstrated. The mAP50 compared with YOLOv5m, YOLOv8m, and YOLOv10m are 0.8\%, 1.2\%, and 3.9\% higher, respectively. The results of the comparative experiments are shown in Table 1.

\begin{table}[!htbp]
    \caption{Comparing the experimental results, the ups and downs of EFA-YOLO on each evaluation metrics are based on the model comparison of the worst case among all models.}
    \centering
    \begin{tabular}{ccccccccc}
        \toprule[1.5pt]
Model & P  & R & mAP & mAP & Params & GFLOPs & Model  & Inference  \\
&(\%)& (\%)&50(\%)&50-95(\%)&(M)&(MB)&Size&Time\\
        \midrule[1pt]

YOLOv5m & 64.8 & 59.3 & 65.4 & 34.5 & 25.0 & 64.0 & 50.5 & 106.2 \\

YOLOv8m & 67.8 & \textbf{\textbf{64.0}} & 65.0 & 33.5 & 25.8 & 78.7 & 49.7 & 123.38 \\

YOLOv9m & \textbf{\textbf{68.7}} & 59.2 & \textbf{\textbf{67.3}} & \textbf{\textbf{36.7}} & 20.0 & 76.5 & 40.8 & 182.36 \\

YOLOv10m & 63.6 & 61.8 & 62.1 & 33.7 & 16.4 & 63.4 & 33.5 & 105.52 \\

EFA-YOLO & 65.3  & 62.8  & 66.2  & 34.1  & \textbf{\textbf{1.4}} & \textbf{\textbf{4.6}} & \textbf{\textbf{3.3}} & \textbf{\textbf{22.19 }} \\
&\textbf{(↑1.7)}&\textbf{(↑3.6)}&\textbf{(↑4.1)}&\textbf{(↑0.6)}&\textbf{(↓24.4)}&\textbf{(↓74.1)}&\textbf{(↓47.2)}&\textbf{(↓160.17)}\\

        \bottomrule[1.5pt]
    \end{tabular}
\end{table}

\begin{itemize}

\item \textbf{Precision (P) and Recall (R) analysis:} In both Precision and Recall, the EFA-YOLO shows significant improvements in both metrics. Compared to YOLOv5m, EFA-YOLO improves Precision by 0.5\% and Recall by 3.5\%. This means that EFA-YOLO outperforms mainstream models such as YOLOv5m and YOLOv8m both in terms of reducing false alarms (i.e., improving Precision) and improving detection comprehensiveness (i.e., improving Recall). Compared with YOLOv10m, EFA-YOLO's Precision and Recall are 1.7\% and 1.0\% higher, respectively, further illustrating the model's improvement in detection accuracy and comprehensiveness.

\item \textbf{mAP50 and mAP50-95 analysis:} The mAP50 is an important index to evaluate the detection accuracy of the model. The experimental results show that the mAP50 of EFA-YOLO is improved by 0.8\% over YOLOv5m, 1.2\% over YOLOv8m, and 4.1\% over YOLOv10m. This significant improvement indicates that EFA-YOLO is more capable of object detection in complex scenarios, especially in the flame detection task, where its high-precision detection advantage is fully realized. In addition, EFA-YOLO performs equally well in the mAP50-95 metrics. Although the improvement is small, the small increase in this metric means that EFA-YOLO is able to maintain high detection accuracy over a wider range of IoU thresholds.

\item \textbf{Params(M), GFLOPs and Model Size analysis:} EFA-YOLO performs particularly well in terms of the number of model parameters (Params) and computational complexity (GFLOPs). Compared to YOLOv5m, YOLOv8m, and YOLOv9m, EFA-YOLO's Parameters are reduced by 94.6\% to only 1.4M, and its GFLOPs are also drastically reduced to 4.6 In terms of model size, EFA-YOLO's Model Size is only 3.3MB, which is 47.2MB less compared to YOLOv5m.

\end{itemize}

\subsection{Detection Result}

\begin{figure}[h]
    \centering
    \includegraphics[width=1\linewidth]{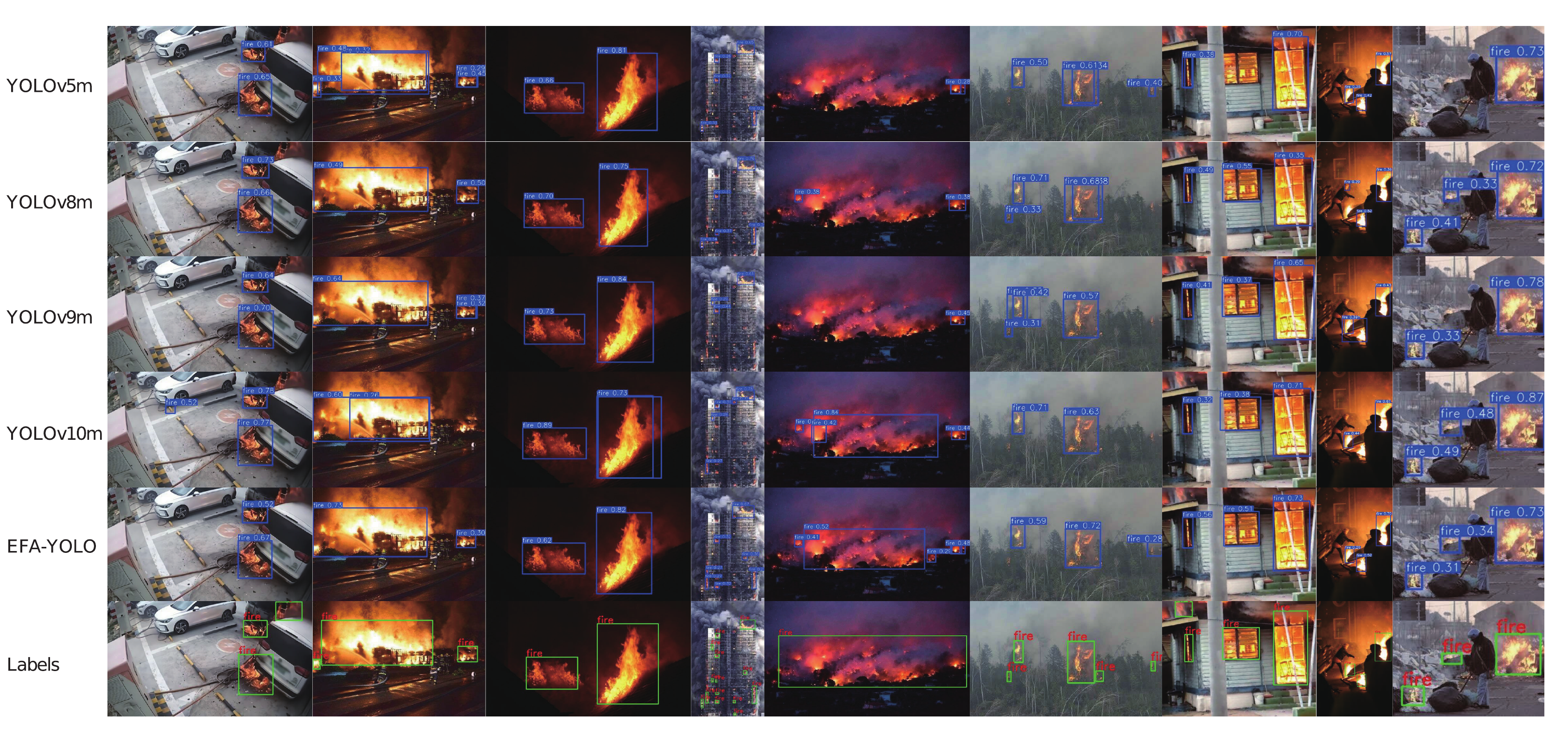}
    \caption{Compare the detection results of the experimental models.}
    \label{fig5}
\end{figure}

\begin{itemize}

\item \textbf{Stability of flame targeting and detection frames:} As can be seen from the Fig~\ref{fig5}, the detection frame of EFA-YOLO is closer to the real label (green frame), especially in the flame detection scene of small targets in complex background, EFA-YOLO can recognize the flame region more accurately, and the bounding box is more compact, which matches the actual flame region more closely. For example, in the flame scene next to the car on the left, both YOLOv5m and YOLOv8m detected the small flame target, but their detection frames showed some deviation and failed to accurately cover the flame region; whereas EFA-YOLO more accurately calibrated the flame, which shows its advantage in handling small targets. In addition, YOLOv9m and YOLOv10m performed well in some scenes, but for situations with complex or heavily occluded backgrounds, the detection frames tended to be on the large side, resulting in the flame target being detected as a larger area and failing to effectively reject background noise. EFA-YOLO, on the other hand, due to the combination of the EAConv and EADown modules, enhances the model's ability to focus on the flame region through the attention mechanism, which can effectively avoid false alarms and omissions, especially in scenarios with complex backgrounds.

\item \textbf{Confidence Score analysis:} From the confidence scores of each model, the detection confidence of EFA-YOLO is generally high, especially when the flame target is small and the boundary of the flame region is not obvious, EFA-YOLO is still able to maintain a high confidence. Meanwhile, under certain extreme lighting conditions, EFA-YOLO is able to maintain high detection confidence, while the confidence of other YOLO models decreases. This indicates that the attention mechanism of EFA-YOLO can effectively enhance the attention to the flame target and improve the robustness of the model when dealing with complex lighting conditions.

\item \textbf{Treatment of occlusion and smoke backgrounds:} EFA-YOLO performs well in handling occlusion and smoke scenes. The last column in the figure shows a large smoke background in the fire scene, and the detection results of other models (e.g., YOLOv9m and YOLOv10m) in the smoke environment are interfered with by misclassifying part of the smoke as a flame target, which results in a large detection frame. In contrast, EFA-YOLO benefits from feature downsampling by the EADown module through the spatial and channel attention mechanism, which can better exclude the interference of background noise and accurately detect the real position of the flame.

\item \textbf{Multi-object detection capability:} EFA-YOLO is more capable of detecting multiple flame targets. In contrast, YOLOv5m and YOLOv8m only detected part of the flame targets with some deviation in the size and position of the detection frame, while EFA-YOLO not only recognized all the flame targets, but also calibrated each flame region more accurately. In the task of multi-object detection, EFA-YOLO demonstrates a clear advantage, showing stronger adaptability to small targets and complex backgrounds.
5. Comparison with real labels: Comparing the detection results of each model with the real label (green box), it can be found that the bounding box generated by EFA-YOLO is closest to the real label, and EFA-YOLO shows higher accuracy and stability, both in the detection of small flame targets and in the detection of targets in complex scenes. This result indicates that EFA-YOLO has a stronger generalization ability under different conditions and is able to stably output high-quality detection results in different scenarios.

\end{itemize}

\subsection{Ablation Study}

In this ablation experiment, we performed a detailed analysis of the performance metrics of the benchmark model YOLOv8m by gradually introducing the EAConv and EADown modules, and evaluated the impact of these modules on the model performance. The experimental results are shown in Table 2, demonstrating the performance of the model in a number of metrics such as Precision (P), Recall (R), mAP50, mAP50-95, Model Parameters (M), Computational Complexity (GFLOPs), and Model Size (MB). The results of the ablation study are shown in Table 2.

\begin{table}[!htbp]
    \caption{ Ablation study results, all model indicators rise and fall according to the benchmark model evaluation metrics.}
    \centering
    \begin{tabular}{cccccc}
        \toprule[1.5pt]
EAConv & EADown & P (\%) & R(\%) & mAP50 (\%) & Map50-95(\%) \\
        \midrule[1pt]

  &   & 64.8 & 59.3 & 65.4 & 34.5 \\

$\checkmark$ &   & \textbf{\textbf{73.4 (↑8.6)}} & 56.7 (↓2.6) & 65.6 \textbf{\textbf{(↑0.2)}} & 34.7 \textbf{\textbf{(↑0.2)}} \\

  & $\checkmark$ & 66.5 \textbf{\textbf{(↑0.7)}} & 57.1 (↓2.2) & 64.7 (↓0.7) & \textbf{\textbf{35.8 (↑1.3)}} \\

$\checkmark$ & $\checkmark$ & 65.3 \textbf{\textbf{(↑0.5)}} & \textbf{\textbf{62.8 (↑3.5)}} & \textbf{\textbf{66.2 (↑0.8)}} & 34.1 (↓0.4)\\
        \midrule[0pt]
        \toprule[1.5pt]

EAConv & EADown & Params(M) & GFLOPs & Model Size(MB) & Inference Time \\
        \midrule[1pt]
  &   & 25 & 64 & 50.5 & 123.38 \\

$\checkmark$ &   & 1.8 \textbf{\textbf{(↓23.2)}} & 5.4 \textbf{\textbf{(↓58.6)}} & 4.0 \textbf{\textbf{(↓46.5)}} & \textbf{\textbf{16.98 (↓106.4)}} \\

  & $\checkmark$ & 2.6 \textbf{\textbf{(↓22.4)}} & 7.5 \textbf{\textbf{(↓56.5)}} & 5.5 \textbf{\textbf{(↓45.0)}} & 25.05 \textbf{\textbf{(↓98.33)}} \\

$\checkmark$ & $\checkmark$ & \textbf{\textbf{1.4 (↓23.6)}} & \textbf{\textbf{4.6 (↓59.4)}} & \textbf{\textbf{3.3 (↓47.2)}} & 22.19 \textbf{\textbf{(↓101.19)}} \\
        \bottomrule[1.5pt]
    \end{tabular}
\end{table}

\begin{itemize}

\item \textbf{Introduce performance analysis for the EAConv module only:} When only the EAConv module was introduced, the model's Precision was significantly improved by 8.6\%, which indicates that the EAConv module improves the model's accuracy in detecting positive categories through effective feature extraction and attention mechanisms. However, Recall decreases, by 2.6\%, which indicates that the model slightly sacrifices in detecting comprehensiveness, possibly due to the attention mechanism focusing on local key features, resulting in some of the edges or non-significant flame targets not being captured. In terms of mAP50 and mAP50-95, the EAConv module has less impact on the detection accuracy of the model, with mAP50 increasing by only 0.2\%, while mAP50-95 slightly improves by 0.2\%, indicating that the module's detection accuracy is more stable at specific thresholds. In terms of model complexity, the EAConv module reduces the amount of parameters of the model by 92.8\% from 25M to 1.8M, which proves that EAConv greatly optimizes the lightness of the model while maintaining a high detection accuracy. The GFLOPs are drastically reduced by 91.6\%, and the computational complexity is significantly reduced. The model size is also reduced from 50.5MB to 4.0MB, a decrease of 46.5MB, which proves that the EAConv module greatly contributes to the lightweighting of the model.

\item I\textbf{ntroduction of performance analysis for the EADown module only:} With the introduction of the EADown module only, the model's Precision and Recall increased by 0.7\% and decreased by 2.2\%, respectively. Although the Precision is increased, the Recall is slightly decreased, indicating that the EADown module may slightly affect the model's ability to capture the full range of targets while the Precision is increased, especially when detecting some flame targets with less detailed features, the model may have missed detections. The accuracy of the model decreases by 0.7\% on mAP50, while it improves on mAP50-95 (increasing by 1.3\%). This indicates that the EADown module, while not contributing much to the overall detection accuracy at specific IoU thresholds (e.g., 0.5), shows better performance in wider IoU intervals (e.g., mAP50-95) and can better cope with the detection of flame targets at different scales. The EADown module likewise performs well in terms of model lightweighting. The amount of parameters in the model is reduced by 89.6\% from 25M to 2.6M, while the GFLOPs are reduced by 88.3\%. The model size was reduced by 45MB from 50.5MB to 5.5MB, indicating that EADown plays an important role in optimizing computational efficiency and reducing model storage requirements.

\item \textbf{Performance analysis of the EAConv and EADown modules was introduced simultaneously:} When both the EAConv and EADown modules were introduced, the model demonstrated overall performance improvement and optimization. Precision increased by 0.5\% to 65.3\%, indicating that the combination of the two performed well in ensuring detection accuracy. Recall increased by 3.5\% to 62.8\%, which is a significant increase in recall compared to when a single module was introduced, indicating that the that the combination is capable of detecting flame targets more comprehensively. On mAP50, the detection accuracy improved by 0.8\%, while mAP50-95 showed a slight decrease (by 0.4\%). This result indicates that although the model's detection performance decreases at higher IoU thresholds, the model exhibits more robust detection in the medium IoU range of mAP50. In terms of model lightweighting, the number of parameters is further reduced to 1.4M, a decrease of 23.6\%, which indicates that the simultaneous introduction of two modules greatly optimizes the number of parameters of the model. GFLOPs are also significantly reduced by 92.8\%, and the computational complexity is significantly optimized. The model size is reduced to 3.3MB, which means that the model is suitable for resource-constrained devices and enables fast and efficient inference.

\end{itemize}

\begin{figure}[h]
    \centering
    \includegraphics[width=1\linewidth]{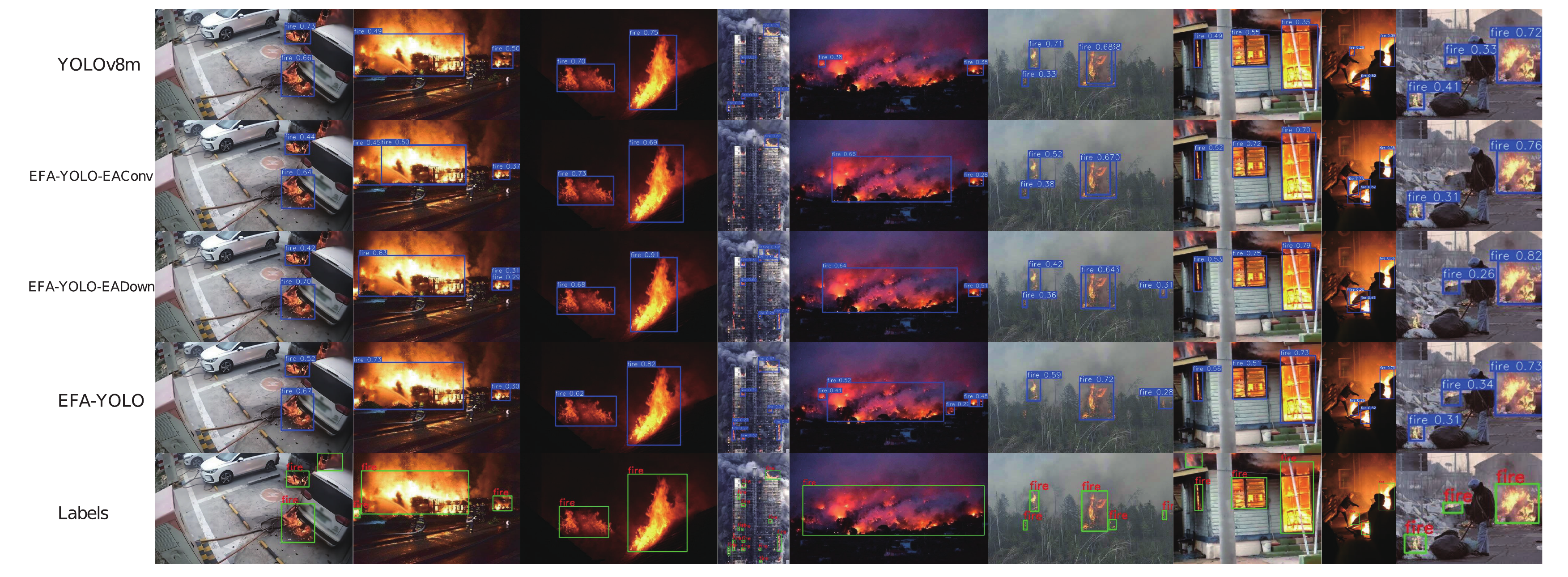}
    \caption{Detection results of the ablation study model. EFA-YOLO-EAConv is EFA-YOLO without the EAConv module, and EFA-YOLO-EADown is EFA-YOLO without the EFA-YOLO-EADown is EFA-YOLO without the EADown module.}
    \label{fig6}
\end{figure}

\begin{itemize}

\item \textbf{Performance of YOLOv8m:} In images, YOLOv8m does not perform well as a benchmark model in flame detection tasks. Although it was able to largely detect flame targets, it showed some limitations in complex backgrounds or small target scenes. For example, in some scenarios (e.g., flames next to a car and a distant fire scene), YOLOv8m fails to accurately capture the actual boundaries of the flames, and the detection frames are significantly larger or smaller compared to the labels of the real flame targets. In addition, the detection confidence of YOLOv8m is low, especially in scenes with long distances or complex backgrounds, and the confidence of some flame targets is only about 0.30, which indicates its poor adaptability to complex environments.

\item \textbf{Removal of the EFA-YOLO (EFA-YOLO-EAConv) manifestation of EAConv:} When the EAConv module is removed, the performance of EFA-YOLO-EAConv is closer to that of YOLOv8m, but with improvements in detail and boundary detection. The absence of the EAConv module results in the model not being able to make full use of the attention mechanism to focus on the critical areas of the flame target during feature extraction, leading to less precise boundaries of the detection box in some scenarios. For example, the flame detection results next to a car show that although the model successfully recognizes the flame target, its bounding box is not compact enough and slightly deviated. Nonetheless, the detection confidence of the model is slightly improved, indicating that EFA-YOLO-EAConv has improved its overall performance, but it is still not optimal.

\item \textbf{Removal of the EFA-YOLO (EFA-YOLO-EADown) manifestation of EADown:} The EFA-YOLO-EADown model is able to capture the features of flame targets better by retaining the EAConv module and removing the EADown module, but shows some limitations when dealing with flame targets at different scales. The absence of the EADown module means that the model loses a part of its optimization of spatial and channel attention during the downsampling process, resulting in a flame boundary detection accuracy decreasing. In the large-scale fire scenario in Fig. EFA-YOLO-EADown fails to accurately capture the full region of the flame, especially in the detection of long-distance flame targets, where the bounding box is significantly smaller and the confidence level is reduced. This suggests that the EADown module plays a key role in multi-scale flame detection, and the lack of this module makes the model's performance less stable when dealing with small flames and long-distance targets.

\item \textbf{Performance of EFA-YOLO in its entirety:} EFA-YOLO combines the EAConv and EADown modules and outperforms YOLOv8m, EFA-YOLO-EAConv and EFA-YOLO-EADown in all scenarios. The model is able to accurately detect the actual area of the flame and the detection frame is highly compatible with the real label (green frame). The performance of EFA-YOLO is very stable in both small target flames and large-scale fire scenarios, and is able to maintain a high confidence level. For example, the detection frame of the flame next to the car closely matches the flame area and has a confidence level of 0.67, which is more accurate compared to other models.

\end{itemize}

\subsection{Error Analysis}

The results of the comparison and ablation experiments are analyzed, and although EFA-YOLO shows excellent performance in flame detection tasks, there are still some errors and limitations. The problem of missed detection mainly appears in small targets and long-distance flame scenes. Although the EADown module optimizes the downsampling process through spatial and channel attention mechanisms, feature loss may still occur in the detection of very small targets, leading to missed detection. In addition, the misdetection problem usually occurs in scenes with complex backgrounds and drastic changes in illumination, and although the EAConv module is able to focus on the flame region through the attention mechanism, the model sometimes still misrecognizes the high-brightness background region as a flame. Bounding box inaccuracy, on the other hand, manifests itself in scenes with blurred boundaries, where the model's bounding box may be large or small when the boundary between the flame and the background is not obvious. Finally, complex environments and occlusion problems remain a major challenge in flame detection, and when the flame is partially occluded, the model may not be able to effectively recognize the flame target in the occluded part.

\subsection{Case Studies}

\begin{itemize}

\item \textbf{Complex background scenes:} In flame detection, complex backgrounds often interfere with the detection results of the model. For example, in a city night scene, the features of the flame and the background are easily confused due to the reflection of lights and the highlighted areas on the surface of buildings. By comparing the results of the experiments, YOLOv5m and YOLOv8m often have false alarms in this kind of scene, and the model recognizes the highlighted areas in the background as flames. On the contrary, EFA-YOLO performs more robustly in such scenarios, and through the attention mechanism of the EAConv module, it is able to effectively distinguish between the flame and the highlighted part in the background, avoiding false alarms. As shown in the figure, EFA-YOLO is able to capture the flame area accurately, and the detection frame highly matches the actual flame area.

\item \textbf{Small object detection:} In the early stage of a fire, the size of the flame is usually small, and the detection performance of traditional detection models in such scenarios is often poor. In this case, EFA-YOLO shows significant advantages in small flame object detection. The ablation experiment shows that after removing the EADown module, the model exhibits inadequate detection in small object detection. In contrast, the complete EFA-YOLO model, with the optimization of the feature downsampling process by the EADown module, effectively retains the key information of small flames, thus improving the detection accuracy of small targets. In contrast, the detection frames of YOLOv8m and YOLOv9m tend to be large or small with low confidence, whereas EFA-YOLO not only accurately detects small flame targets with high confidence.

\item \textbf{Sheltered scenes:} In some fire scenarios, the flame target may be partially occluded (e.g., trees, buildings, etc.), which poses a greater challenge to the detection task. Conventional object detection models often have difficulty accurately detecting the occluded flame portion in such scenarios, and the detection frame usually covers only the visible flame portion and misses the occluded region. However, EFA-YOLO is able to infer the flame region of the occluded part through the contextual information through the global attention mechanism of the EAConv module, thus ensuring the integrity of the detection results. In the comparison experiments, EFA-YOLO successfully detects the partially occluded flame region, while YOLOv8m and YOLOv9m show obvious missed detection in the same scene.

\item \textbf{Remote flame detection:} Long-range fire detection is a key application scenario in fire monitoring. When the flame target is far away from the camera, the resolution of the flame in the image is low, and the traditional model is prone to miss or misdetect it. In this case, EFA-YOLO is able to maintain high detection accuracy in long-distance scenarios thanks to its multi-scale feature extraction and attention mechanism. In the comparison experiments, YOLOv8m and YOLOv9m have low confidence in long-distance flame object detection, and some flame targets are not even detected, while EFA-YOLO is able to successfully recognize these long-distance flame targets, and the detection frame is highly consistent with the real label.

\end{itemize}

\section{Conclusion}

In this paper, an efficient lightweight model EFA-YOLO for fire flame detection is proposed, which realizes efficient feature extraction and downsampling by introducing two key modules, EAConv and EADown, and thus improves the detection accuracy, computational efficiency and real-time performance in the flame detection task. The experimental results show that EFA-YOLO performs well in terms of the number of model parameters, computational complexity and inference speed, and the number of model parameters is reduced by 94.6\% and the inference time is improved by 88 times compared with the mainstream YOLO series models, which significantly improves the model's lightweight and real-time performance. Although EFA-YOLO shows good performance in fire detection tasks, there is still some room for improvement. First, although the model strikes a good balance between detection accuracy and computational efficiency, it may still suffer from missed or false detections when dealing with extremely dense scenes. This is mainly due to the fact that the model still has some limitations when dealing with very small-scale flame targets or complex backgrounds (e.g., smoke, light reflections). Future work will optimize for this limitation by introducing local information awareness to enhance the detection of dense small targets, further optimizing the multi-scale feature fusion strategy, and introducing timing information to improve the model's performance in video detection. Better technical support is provided for real-time fire detection and complex scene adaptation.

\bibliography{sn-bibliography}


\begin{thebibliography}{34}
\ifx \bisbn   \undefined \def \bisbn  #1{ISBN #1}\fi
\ifx \binits  \undefined \def \binits#1{#1}\fi
\ifx \bauthor  \undefined \def \bauthor#1{#1}\fi
\ifx \batitle  \undefined \def \batitle#1{#1}\fi
\ifx \bjtitle  \undefined \def \bjtitle#1{#1}\fi
\ifx \bvolume  \undefined \def \bvolume#1{\textbf{#1}}\fi
\ifx \byear  \undefined \def \byear#1{#1}\fi
\ifx \bissue  \undefined \def \bissue#1{#1}\fi
\ifx \bfpage  \undefined \def \bfpage#1{#1}\fi
\ifx \blpage  \undefined \def \blpage #1{#1}\fi
\ifx \burl  \undefined \def \burl#1{\textsf{#1}}\fi
\ifx \doiurl  \undefined \def \doiurl#1{\url{https://doi.org/#1}}\fi
\ifx \betal  \undefined \def \betal{\textit{et al.}}\fi
\ifx \binstitute  \undefined \def \binstitute#1{#1}\fi
\ifx \binstitutionaled  \undefined \def \binstitutionaled#1{#1}\fi
\ifx \bctitle  \undefined \def \bctitle#1{#1}\fi
\ifx \beditor  \undefined \def \beditor#1{#1}\fi
\ifx \bpublisher  \undefined \def \bpublisher#1{#1}\fi
\ifx \bbtitle  \undefined \def \bbtitle#1{#1}\fi
\ifx \bedition  \undefined \def \bedition#1{#1}\fi
\ifx \bseriesno  \undefined \def \bseriesno#1{#1}\fi
\ifx \blocation  \undefined \def \blocation#1{#1}\fi
\ifx \bsertitle  \undefined \def \bsertitle#1{#1}\fi
\ifx \bsnm \undefined \def \bsnm#1{#1}\fi
\ifx \bsuffix \undefined \def \bsuffix#1{#1}\fi
\ifx \bparticle \undefined \def \bparticle#1{#1}\fi
\ifx \barticle \undefined \def \barticle#1{#1}\fi
\bibcommenthead
\ifx \bconfdate \undefined \def \bconfdate #1{#1}\fi
\ifx \botherref \undefined \def \botherref #1{#1}\fi
\ifx \url \undefined \def \url#1{\textsf{#1}}\fi
\ifx \bchapter \undefined \def \bchapter#1{#1}\fi
\ifx \bbook \undefined \def \bbook#1{#1}\fi
\ifx \bcomment \undefined \def \bcomment#1{#1}\fi
\ifx \oauthor \undefined \def \oauthor#1{#1}\fi
\ifx \citeauthoryear \undefined \def \citeauthoryear#1{#1}\fi
\ifx \endbibitem  \undefined \def \endbibitem {}\fi
\ifx \bconflocation  \undefined \def \bconflocation#1{#1}\fi
\ifx \arxivurl  \undefined \def \arxivurl#1{\textsf{#1}}\fi
\csname PreBibitemsHook\endcsname

\bibitem[\protect\citeauthoryear{Jones et~al.}{2024}]{1}
\begin{barticle}
\bauthor{\bsnm{Jones}, \binits{M.W.}},
\bauthor{\bsnm{Kelley}, \binits{D.I.}},
\bauthor{\bsnm{Burton}, \binits{C.A.}},
\bauthor{\bsnm{Di~Giuseppe}, \binits{F.}},
\bauthor{\bsnm{Barbosa}, \binits{M.L.F.}},
\bauthor{\bsnm{Brambleby}, \binits{E.}},
\bauthor{\bsnm{Hartley}, \binits{A.J.}},
\bauthor{\bsnm{Lombardi}, \binits{A.}},
\bauthor{\bsnm{Mataveli}, \binits{G.}},
\bauthor{\bsnm{McNorton}, \binits{J.R.}},
\bauthor{\bsnm{Spuler}, \binits{F.R.}},
\bauthor{\bsnm{Wessel}, \binits{J.B.}},
\bauthor{\bsnm{Abatzoglou}, \binits{J.T.}},
\bauthor{\bsnm{Anderson}, \binits{L.O.}},
\bauthor{\bsnm{Andela}, \binits{N.}},
\bauthor{\bsnm{Archibald}, \binits{S.}},
\bauthor{\bsnm{Armenteras}, \binits{D.}},
\bauthor{\bsnm{Burke}, \binits{E.}},
\bauthor{\bsnm{Carmenta}, \binits{R.}},
\bauthor{\bsnm{Chuvieco}, \binits{E.}},
\bauthor{\bsnm{Clarke}, \binits{H.}},
\bauthor{\bsnm{Doerr}, \binits{S.H.}},
\bauthor{\bsnm{Fernandes}, \binits{P.M.}},
\bauthor{\bsnm{Giglio}, \binits{L.}},
\bauthor{\bsnm{Hamilton}, \binits{D.S.}},
\bauthor{\bsnm{Hantson}, \binits{S.}},
\bauthor{\bsnm{Harris}, \binits{S.}},
\bauthor{\bsnm{Jain}, \binits{P.}},
\bauthor{\bsnm{Kolden}, \binits{C.A.}},
\bauthor{\bsnm{Kurvits}, \binits{T.}},
\bauthor{\bsnm{Lampe}, \binits{S.}},
\bauthor{\bsnm{Meier}, \binits{S.}},
\bauthor{\bsnm{New}, \binits{S.}},
\bauthor{\bsnm{Parrington}, \binits{M.}},
\bauthor{\bsnm{Perron}, \binits{M.M.G.}},
\bauthor{\bsnm{Qu}, \binits{Y.}},
\bauthor{\bsnm{Ribeiro}, \binits{N.S.}},
\bauthor{\bsnm{Saharjo}, \binits{B.H.}},
\bauthor{\bsnm{San-Miguel-Ayanz}, \binits{J.}},
\bauthor{\bsnm{Shuman}, \binits{J.K.}},
\bauthor{\bsnm{Tanpipat}, \binits{V.}},
\bauthor{\bsnm{Werf}, \binits{G.R.}},
\bauthor{\bsnm{Veraverbeke}, \binits{S.}},
\bauthor{\bsnm{Xanthopoulos}, \binits{G.}}:
\batitle{State of wildfires 2023--2024}.
\bjtitle{Earth System Science Data}
\bvolume{16}(\bissue{8}),
\bfpage{3601}--\blpage{3685}
(\byear{2024})
\doiurl{10.5194/essd-16-3601-2024}
\end{barticle}
\endbibitem

\bibitem[\protect\citeauthoryear{Silva et~al.}{2024}]{2}
\begin{barticle}
\bauthor{\bsnm{Silva}, \binits{D.}},
\bauthor{\bsnm{Ferreira}, \binits{T.M.}},
\bauthor{\bsnm{Rodrigues}, \binits{H.}}:
\batitle{Wildland-urban interface fire exposure of rural settlements: The case
  of montesinho natural park}.
\bjtitle{International Journal of Disaster Risk Reduction}
\bvolume{112},
\bfpage{104790}
(\byear{2024})
\doiurl{10.1016/j.ijdrr.2024.104790}
\end{barticle}
\endbibitem

\bibitem[\protect\citeauthoryear{Santos et~al.}{2024}]{3}
\begin{barticle}
\bauthor{\bsnm{Santos}, \binits{R.}},
\bauthor{\bsnm{Russo}, \binits{A.}},
\bauthor{\bsnm{Gouveia}, \binits{C.M.}}:
\batitle{Co-occurrence of marine and atmospheric heatwaves with drought
  conditions and fire activity in the mediterranean region}.
\bjtitle{Scientific Reports}
\bvolume{14}(\bissue{1}),
\bfpage{19233}
(\byear{2024})
\doiurl{10.1038/s41598-024-69691-y}
\end{barticle}
\endbibitem

\bibitem[\protect\citeauthoryear{Shiroshita et~al.}{2024}]{4}
\begin{barticle}
\bauthor{\bsnm{Shiroshita}, \binits{H.}},
\bauthor{\bsnm{Jayaratne}, \binits{R.}},
\bauthor{\bsnm{Kitagawa}, \binits{K.}}:
\batitle{Integrating communities’ perspectives in understanding disaster
  risk}.
\bjtitle{Natural Hazards}
\bvolume{120}(\bissue{9}),
\bfpage{8263}--\blpage{8282}
(\byear{2024})
\doiurl{10.1007/s11069-024-06452-0}
\end{barticle}
\endbibitem

\bibitem[\protect\citeauthoryear{Sankey et~al.}{2024}]{5}
\begin{barticle}
\bauthor{\bsnm{Sankey}, \binits{T.T.}},
\bauthor{\bsnm{Tango}, \binits{L.}},
\bauthor{\bsnm{Tatum}, \binits{J.}},
\bauthor{\bsnm{Sankey}, \binits{J.B.}}:
\batitle{Forest fire, thinning, and flood in wildland-urban interface: Uav and
  lidar-based estimate of natural disaster impacts}.
\bjtitle{Landscape Ecology}
\bvolume{39}(\bissue{3}),
\bfpage{58}
(\byear{2024})
\doiurl{10.1007/s10980-024-01811-5}
\end{barticle}
\endbibitem

\bibitem[\protect\citeauthoryear{Celic et~al.}{2024}]{6}
\begin{barticle}
\bauthor{\bsnm{Celic}, \binits{B.}},
\bauthor{\bsnm{Kieseberg}, \binits{K.}},
\bauthor{\bsnm{Garn}, \binits{B.}},
\bauthor{\bsnm{Simos}, \binits{D.E.}}:
\batitle{Disaster incident analysis via algebra stories}.
\bjtitle{Mathematics in Computer Science}
\bvolume{18}(\bissue{2}),
\bfpage{11}
(\byear{2024})
\doiurl{10.1007/s11786-024-00586-x}
\end{barticle}
\endbibitem

\bibitem[\protect\citeauthoryear{kas kozani et~al.}{2025}]{7}
\begin{barticle}
\bauthor{\bsnm{kozani}, \binits{s.}},
\bauthor{\bsnm{yousefiniya}, \binits{m.}},
\bauthor{\bsnm{einodin}, \binits{h.}},
\bauthor{\bsnm{nejad}, \binits{h.}}:
\batitle{Analysis of the spatial structure of places prone to accidents (fires)
  in the city using the capabilities of arc gis software}.
\bjtitle{Journal of Environmental Science Studies}
\bvolume{9}(\bissue{4}),
\bfpage{9490}--\blpage{9480}
(\byear{2025})
\doiurl{10.22034/jess.2024.448346.2242}
{\href{https://arxiv.org/abs/https://www.jess.ir/article 198956
  62b2b69a1e00d50ad8aa507bf668743a.pdf}{{https://www.jess.ir/article 198956
  62b2b69a1e00d50ad8aa507bf668743a.pdf}}}
\end{barticle}
\endbibitem

\bibitem[\protect\citeauthoryear{Chen et~al.}{2024}]{8}
\begin{barticle}
\bauthor{\bsnm{Chen}, \binits{Y.}},
\bauthor{\bsnm{Jiang}, \binits{Y.}},
\bauthor{\bsnm{Xu}, \binits{Z.-d.}},
\bauthor{\bsnm{Zhang}, \binits{L.}},
\bauthor{\bsnm{Yan}, \binits{F.}},
\bauthor{\bsnm{Zong}, \binits{H.}}:
\batitle{A lightweight fire hazard recognition model for urban subterranean
  buildings suitable for resource-constrained embedded systems}.
\bjtitle{Signal, Image and Video Processing}
\bvolume{18}(\bissue{10}),
\bfpage{6645}--\blpage{6659}
(\byear{2024})
\doiurl{10.1007/s11760-024-03341-8}
\end{barticle}
\endbibitem

\bibitem[\protect\citeauthoryear{Meng et~al.}{2024}]{9}
\begin{barticle}
\bauthor{\bsnm{Meng}, \binits{L.}},
\bauthor{\bsnm{O’Hehir}, \binits{J.}},
\bauthor{\bsnm{Gao}, \binits{J.}},
\bauthor{\bsnm{Peters}, \binits{S.}},
\bauthor{\bsnm{Hay}, \binits{A.}}:
\batitle{A theoretical framework for improved fire suppression by linking
  management models with smart early fire detection and suppression
  technologies}.
\bjtitle{Journal of Forestry Research}
\bvolume{35}(\bissue{1}),
\bfpage{86}
(\byear{2024})
\doiurl{10.1007/s11676-024-01737-3}
\end{barticle}
\endbibitem

\bibitem[\protect\citeauthoryear{Daoud et~al.}{2024}]{10}
\begin{barticle}
\bauthor{\bsnm{Daoud}, \binits{Z.}},
\bauthor{\bsnm{Hamida}, \binits{A.B.}},
\bauthor{\bsnm{Amar}, \binits{C.B.}},
\bauthor{\bsnm{Miguet}, \binits{S.}}:
\batitle{A one stream three-dimensional convolutional neural network for fire
  recognition based on spatio-temporal fire analysis}.
\bjtitle{Evolving Systems}
(\byear{2024})
\doiurl{10.1007/s12530-024-09623-3}
\end{barticle}
\endbibitem

\bibitem[\protect\citeauthoryear{Özel et~al.}{2024}]{11}
\begin{botherref}
\oauthor{\bsnm{Özel}, \binits{B.}},
\oauthor{\bsnm{Alam}, \binits{M.S.}},
\oauthor{\bsnm{Khan}, \binits{M.U.}}:
Review of modern forest fire detection techniques: Innovations in image
  processing and deep learning.
Information
\textbf{15}(9)
(2024)
\doiurl{10.3390/info15090538}
\end{botherref}
\endbibitem

\bibitem[\protect\citeauthoryear{Shawly and Alsheikhy}{2024}]{12}
\begin{barticle}
\bauthor{\bsnm{Shawly}, \binits{T.}},
\bauthor{\bsnm{Alsheikhy}, \binits{A.A.}}:
\batitle{Fire identification based on novel dense generative adversarial
  networks}.
\bjtitle{Artificial Intelligence Review}
\bvolume{57}(\bissue{8}),
\bfpage{207}
(\byear{2024})
\doiurl{10.1007/s10462-024-10848-6}
\end{barticle}
\endbibitem

\bibitem[\protect\citeauthoryear{Wang et~al.}{2024}]{13}
\begin{barticle}
\bauthor{\bsnm{Wang}, \binits{J.}},
\bauthor{\bsnm{Wang}, \binits{C.}},
\bauthor{\bsnm{Ding}, \binits{W.}},
\bauthor{\bsnm{Li}, \binits{C.}}:
\batitle{{YOLOv5s-ACE}: Forest fire object detection algorithm based on
  improved yolov5s}.
\bjtitle{Fire Technology}
(\byear{2024})
\doiurl{10.1007/s10694-024-01619-4}
\end{barticle}
\endbibitem

\bibitem[\protect\citeauthoryear{Sun et~al.}{2023}]{14}
\begin{barticle}
\bauthor{\bsnm{Sun}, \binits{B.}},
\bauthor{\bsnm{Wang}, \binits{Y.}},
\bauthor{\bsnm{Wu}, \binits{S.}}:
\batitle{An efficient lightweight cnn model for real-time fire smoke
  detection}.
\bjtitle{Journal of Real-Time Image Processing}
\bvolume{20}(\bissue{4}),
\bfpage{74}
(\byear{2023})
\doiurl{10.1007/s11554-023-01331-6}
\end{barticle}
\endbibitem

\bibitem[\protect\citeauthoryear{Cao et~al.}{2024}]{15}
\begin{barticle}
\bauthor{\bsnm{Cao}, \binits{L.}},
\bauthor{\bsnm{Shen}, \binits{Z.}},
\bauthor{\bsnm{Xu}, \binits{S.}}:
\batitle{Efficient forest fire detection based on an improved yolo model}.
\bjtitle{Visual Intelligence}
\bvolume{2}(\bissue{1}),
\bfpage{20}
(\byear{2024})
\doiurl{10.1007/s44267-024-00053-y}
\end{barticle}
\endbibitem

\bibitem[\protect\citeauthoryear{Yang et~al.}{2024}]{16}
\begin{barticle}
\bauthor{\bsnm{Yang}, \binits{W.}},
\bauthor{\bsnm{Wu}, \binits{Y.}},
\bauthor{\bsnm{Chow}, \binits{S.K.K.}}:
\batitle{Deep learning method for real-time fire detection system for urban
  fire monitoring and control}.
\bjtitle{International Journal of Computational Intelligence Systems}
\bvolume{17}(\bissue{1}),
\bfpage{216}
(\byear{2024})
\doiurl{10.1007/s44196-024-00592-8}
\end{barticle}
\endbibitem

\bibitem[\protect\citeauthoryear{Yu and Kim}{2024}]{17}
\begin{barticle}
\bauthor{\bsnm{Yu}, \binits{R.}},
\bauthor{\bsnm{Kim}, \binits{K.}}:
\batitle{A study of novel initial fire detection algorithm based on deep
  learning method}.
\bjtitle{Journal of Electrical Engineering \& Technology}
\bvolume{19}(\bissue{6}),
\bfpage{3675}--\blpage{3686}
(\byear{2024})
\doiurl{10.1007/s42835-024-02009-0}
\end{barticle}
\endbibitem

\bibitem[\protect\citeauthoryear{Lv et~al.}{2024}]{18}
\begin{barticle}
\bauthor{\bsnm{Lv}, \binits{C.}},
\bauthor{\bsnm{Zhou}, \binits{H.}},
\bauthor{\bsnm{Chen}, \binits{Y.}},
\bauthor{\bsnm{Fan}, \binits{D.}},
\bauthor{\bsnm{Di}, \binits{F.}}:
\batitle{A lightweight fire detection algorithm for small targets based on
  yolov5s}.
\bjtitle{Scientific Reports}
\bvolume{14}(\bissue{1}),
\bfpage{14104}
(\byear{2024})
\doiurl{10.1038/s41598-024-64934-4}
\end{barticle}
\endbibitem

\bibitem[\protect\citeauthoryear{Wang et~al.}{2024}]{19}
\begin{barticle}
\bauthor{\bsnm{Wang}, \binits{Z.}},
\bauthor{\bsnm{Zhao}, \binits{X.}},
\bauthor{\bsnm{Li}, \binits{D.}}:
\batitle{Datfnets-dynamic adaptive assigned transformer network for fire
  detection}.
\bjtitle{Complex \& Intelligent Systems}
\bvolume{10}(\bissue{4}),
\bfpage{5703}--\blpage{5720}
(\byear{2024})
\doiurl{10.1007/s40747-024-01444-w}
\end{barticle}
\endbibitem

\bibitem[\protect\citeauthoryear{Cao et~al.}{2024}]{20}
\begin{barticle}
\bauthor{\bsnm{Cao}, \binits{X.}},
\bauthor{\bsnm{Wu}, \binits{J.}},
\bauthor{\bsnm{Chen}, \binits{J.}},
\bauthor{\bsnm{Li}, \binits{Z.}}:
\batitle{Complex scenes fire object detection based on feature fusion and
  channel attention}.
\bjtitle{Arabian Journal for Science and Engineering}
(\byear{2024})
\doiurl{10.1007/s13369-024-09471-y}
\end{barticle}
\endbibitem

\bibitem[\protect\citeauthoryear{Chen et~al.}{2024}]{21}
\begin{barticle}
\bauthor{\bsnm{Chen}, \binits{Y.}},
\bauthor{\bsnm{Li}, \binits{J.}},
\bauthor{\bsnm{Sun}, \binits{K.}},
\bauthor{\bsnm{Zhang}, \binits{Y.}}:
\batitle{A lightweight early forest fire and smoke detection method}.
\bjtitle{The Journal of Supercomputing}
\bvolume{80}(\bissue{7}),
\bfpage{9870}--\blpage{9893}
(\byear{2024})
\doiurl{10.1007/s11227-023-05835-7}
\end{barticle}
\endbibitem

\bibitem[\protect\citeauthoryear{Wu et~al.}{2024}]{22}
\begin{barticle}
\bauthor{\bsnm{Wu}, \binits{S.}},
\bauthor{\bsnm{Sheng}, \binits{B.}},
\bauthor{\bsnm{Fu}, \binits{G.}},
\bauthor{\bsnm{Zhang}, \binits{D.}},
\bauthor{\bsnm{Jian}, \binits{Y.}}:
\batitle{Multiscale fire image detection method based on cnn and transformer}.
\bjtitle{Multimedia Tools and Applications}
\bvolume{83}(\bissue{16}),
\bfpage{49787}--\blpage{49811}
(\byear{2024})
\doiurl{10.1007/s11042-023-17482-4}
\end{barticle}
\endbibitem

\bibitem[\protect\citeauthoryear{Li et~al.}{2024}]{23}
\begin{barticle}
\bauthor{\bsnm{Li}, \binits{B.}},
\bauthor{\bsnm{Xu}, \binits{F.}},
\bauthor{\bsnm{Li}, \binits{X.}},
\bauthor{\bsnm{Yu}, \binits{C.}},
\bauthor{\bsnm{Zhang}, \binits{X.}}:
\batitle{Early stage fire detection system based on shallow guide deep
  network}.
\bjtitle{Fire Technology}
\bvolume{60}(\bissue{3}),
\bfpage{1803}--\blpage{1821}
(\byear{2024})
\doiurl{10.1007/s10694-024-01549-1}
\end{barticle}
\endbibitem

\bibitem[\protect\citeauthoryear{Dou et~al.}{2024}]{24}
\begin{barticle}
\bauthor{\bsnm{Dou}, \binits{Z.}},
\bauthor{\bsnm{Zhou}, \binits{H.}},
\bauthor{\bsnm{Liu}, \binits{Z.}},
\bauthor{\bsnm{Hu}, \binits{Y.}},
\bauthor{\bsnm{Wang}, \binits{P.}},
\bauthor{\bsnm{Zhang}, \binits{J.}},
\bauthor{\bsnm{Wang}, \binits{Q.}},
\bauthor{\bsnm{Chen}, \binits{L.}},
\bauthor{\bsnm{Diao}, \binits{X.}},
\bauthor{\bsnm{Li}, \binits{J.}}:
\batitle{An improved yolov5s fire detection model}.
\bjtitle{Fire Technology}
\bvolume{60}(\bissue{1}),
\bfpage{135}--\blpage{166}
(\byear{2024})
\doiurl{10.1007/s10694-023-01492-7}
\end{barticle}
\endbibitem

\bibitem[\protect\citeauthoryear{Wang et~al.}{2024a}]{25}
\begin{barticle}
\bauthor{\bsnm{Wang}, \binits{D.}},
\bauthor{\bsnm{Qian}, \binits{Y.}},
\bauthor{\bsnm{Lu}, \binits{J.}},
\bauthor{\bsnm{Wang}, \binits{P.}},
\bauthor{\bsnm{Hu}, \binits{Z.}},
\bauthor{\bsnm{Chai}, \binits{Y.}}:
\batitle{{Fs-yolo}: fire-smoke detection based on improved yolov7}.
\bjtitle{Multimedia Systems}
\bvolume{30}(\bissue{4}),
\bfpage{215}
(\byear{2024})
\doiurl{10.1007/s00530-024-01359-z}
\end{barticle}
\endbibitem

\bibitem[\protect\citeauthoryear{Wang et~al.}{2024b}]{26}
\begin{barticle}
\bauthor{\bsnm{Wang}, \binits{T.}},
\bauthor{\bsnm{Wang}, \binits{J.}},
\bauthor{\bsnm{Wang}, \binits{C.}},
\bauthor{\bsnm{Lei}, \binits{Y.}},
\bauthor{\bsnm{Cao}, \binits{R.}},
\bauthor{\bsnm{Wang}, \binits{L.}}:
\batitle{Improving yolox network for multi-scale fire detection}.
\bjtitle{The Visual Computer}
\bvolume{40}(\bissue{9}),
\bfpage{6493}--\blpage{6505}
(\byear{2024})
\doiurl{10.1007/s00371-023-03178-1}
\end{barticle}
\endbibitem

\bibitem[\protect\citeauthoryear{Zhang et~al.}{2024}]{27}
\begin{botherref}
\oauthor{\bsnm{Zhang}, \binits{T.}},
\oauthor{\bsnm{Li}, \binits{L.}},
\oauthor{\bsnm{Zhou}, \binits{Y.}},
\oauthor{\bsnm{Liu}, \binits{W.}},
\oauthor{\bsnm{Qian}, \binits{C.}},
\oauthor{\bsnm{Ji}, \binits{X.}}:
CAS-ViT: Convolutional Additive Self-attention Vision Transformers for
  Efficient Mobile Applications
(2024).
\url{https://arxiv.org/abs/2408.03703}
\end{botherref}
\endbibitem

\bibitem[\protect\citeauthoryear{Chollet}{2017}]{28}
\begin{bchapter}
\bauthor{\bsnm{Chollet}, \binits{F.}}:
\bctitle{Xception: Deep learning with depthwise separable convolutions}.
In: \bbtitle{2017 IEEE Conference on Computer Vision and Pattern Recognition
  (CVPR)},
pp. \bfpage{1800}--\blpage{1807}
(\byear{2017}).
\doiurl{10.1109/CVPR.2017.195}
\end{bchapter}
\endbibitem

\bibitem[\protect\citeauthoryear{Geng}{2023}]{29}
\begin{botherref}
\oauthor{\bsnm{Geng}, \binits{Y.}}:
fire-smoke-detect-yolov4: fire-smoke-detect-yolov5 and
  fire-smoke-detection-dataset.
\url{https://github.com/gengyanlei/fire-smoke-detect-yolov4}
(2023)
\end{botherref}
\endbibitem

\bibitem[\protect\citeauthoryear{Chen et~al.}{2024}]{30}
\begin{barticle}
\bauthor{\bsnm{Chen}, \binits{W.}},
\bauthor{\bsnm{Luo}, \binits{J.}},
\bauthor{\bsnm{Zhang}, \binits{F.}},
\bauthor{\bsnm{Tian}, \binits{Z.}}:
\batitle{A review of object detection: Datasets, performance evaluation,
  architecture, applications and current trends}.
\bjtitle{Multimedia Tools and Applications}
\bvolume{83}(\bissue{24}),
\bfpage{65603}--\blpage{65661}
(\byear{2024})
\doiurl{10.1007/s11042-023-17949-4}
\end{barticle}
\endbibitem

\bibitem[\protect\citeauthoryear{Jocher}{2020}]{31}
\begin{botherref}
\oauthor{\bsnm{Jocher}, \binits{G.}}:
{YOLOv5 by Ultralytics}.
\doiurl{10.5281/zenodo.3908559} .
\url{https://github.com/ultralytics/yolov5}
\end{botherref}
\endbibitem

\bibitem[\protect\citeauthoryear{Jocher et~al.}{2023}]{32}
\begin{botherref}
\oauthor{\bsnm{Jocher}, \binits{G.}},
\oauthor{\bsnm{Chaurasia}, \binits{A.}},
\oauthor{\bsnm{Qiu}, \binits{J.}}:
{Ultralytics YOLO}.
\url{https://github.com/ultralytics/ultralytics}
\end{botherref}
\endbibitem

\bibitem[\protect\citeauthoryear{Wang et~al.}{2024a}]{33}
\begin{botherref}
\oauthor{\bsnm{Wang}, \binits{C.-Y.}},
\oauthor{\bsnm{Yeh}, \binits{I.-H.}},
\oauthor{\bsnm{Liao}, \binits{H.-Y.M.}}:
YOLOv9: Learning What You Want to Learn Using Programmable Gradient Information
(2024).
\url{https://arxiv.org/abs/2402.13616}
\end{botherref}
\endbibitem

\bibitem[\protect\citeauthoryear{Wang et~al.}{2024b}]{34}
\begin{botherref}
\oauthor{\bsnm{Wang}, \binits{A.}},
\oauthor{\bsnm{Chen}, \binits{H.}},
\oauthor{\bsnm{Liu}, \binits{L.}},
\oauthor{\bsnm{Chen}, \binits{K.}},
\oauthor{\bsnm{Lin}, \binits{Z.}},
\oauthor{\bsnm{Han}, \binits{J.}},
\oauthor{\bsnm{Ding}, \binits{G.}}:
YOLOv10: Real-Time End-to-End Object Detection
(2024).
\url{https://arxiv.org/abs/2405.14458}
\end{botherref}
\endbibitem

\end{thebibliography}

\end{document}